\documentclass[runningheads]{llncs}
\usepackage{graphicx}

\usepackage{bm}
\usepackage{tikz}
\usepackage{comment}
\usepackage{subcaption}
\usepackage{mathtools}
\usepackage{amsmath,amssymb} 
\usepackage{color}
\usepackage{todonotes}
\usepackage{booktabs}
\usepackage{hyperref}

\newcommand{\ourmethod}[1]{{NeuralSI}}

\begin{document}
\pagestyle{headings}
\mainmatter

\title{\ourmethod{}: Structural Parameter Identification in Nonlinear Dynamical Systems}


\titlerunning{Structural Parameter Identification in Nonlinear Dynamical Systems}

\author{Xuyang Li \and
Hamed Bolandi \and
Talal Salem \and
Nizar Lajnef \and
Vishnu Naresh Boddeti}

\authorrunning{X. Li et al.}
\institute{Michigan State University\\
\email{\{lixuyan1, bolandih, salemtal, lajnefni, vishnu\}@msu.edu}
}

\maketitle

\begin{abstract}
Structural monitoring for complex built environments often suffers from mismatch between design, laboratory testing, and actual built parameters. Additionally, real-world structural identification problems encounter many challenges. For example, the lack of accurate baseline models, high dimensionality, and complex multivariate partial differential equations (PDEs) pose significant difficulties in training and learning conventional data-driven algorithms. This paper explores a new framework, dubbed \ourmethod{}, for structural identification by augmenting PDEs that govern structural dynamics with neural networks. Our approach seeks to estimate nonlinear parameters from governing equations. We consider the vibration of nonlinear beams with two unknown parameters, one that represents geometric and material variations, and another that captures energy losses in the system mainly through damping. The data for parameter estimation is obtained from a limited set of measurements, which is conducive to applications in structural health monitoring where the exact state of an existing structure is typically unknown and only a limited amount of data samples can be collected in the field. The trained model can also be extrapolated under both standard and extreme conditions using the identified structural parameters. We compare with pure data-driven Neural Networks and other classical Physics-Informed Neural Networks (PINNs). Our approach reduces both interpolation and extrapolation errors in displacement distribution by two to five orders of magnitude over the baselines. Code is available at \url{https://github.com/human-analysis/neural-structural-identification}
\end{abstract}

\section{Introduction}
Structural-system identification (SI) \cite{yuen2004two,lai2019sparse,bagheri2018structural,ghorbani2020hybrid,tuhta2019multi,zhou2022semi} refers to methods for inverse calculation of structural systems using data to calibrate a mathematical or digital model. The calibrated models are then used to either estimate or predict the future performance of structural systems and, eventually, their remaining useful life. Non-linear structural systems with spatial and temporal variations present a particular challenge for most inverse identification methods \cite{lai2021structural,banerjee2010self,entezami2017structural}. In dynamic analysis of civil structural systems, prior research efforts primarily focused on matching experimental data with either mechanistic models (i.e., known mechanical models) \cite{rezaiee2020sensitivity,yin2017vibration} or with black box models with only input/output information (i.e., purely data-driven approaches), \cite{sarmadi2020energy,entezami2020big,diez2016clustering}. Examples of these approaches include eigensystem identification algorithms\cite{yang2019automated}, frequency domain decomposition \cite{brincker2001modal}, stochastic optimization techniques \cite{reynders2008reference}, and sparse identification\cite{brunton2016discovering}. A majority of these approaches, however, fail to capture highly non-linear behaviors.
\begin{figure}[t]
    \centering    
    \includegraphics[width=\textwidth]{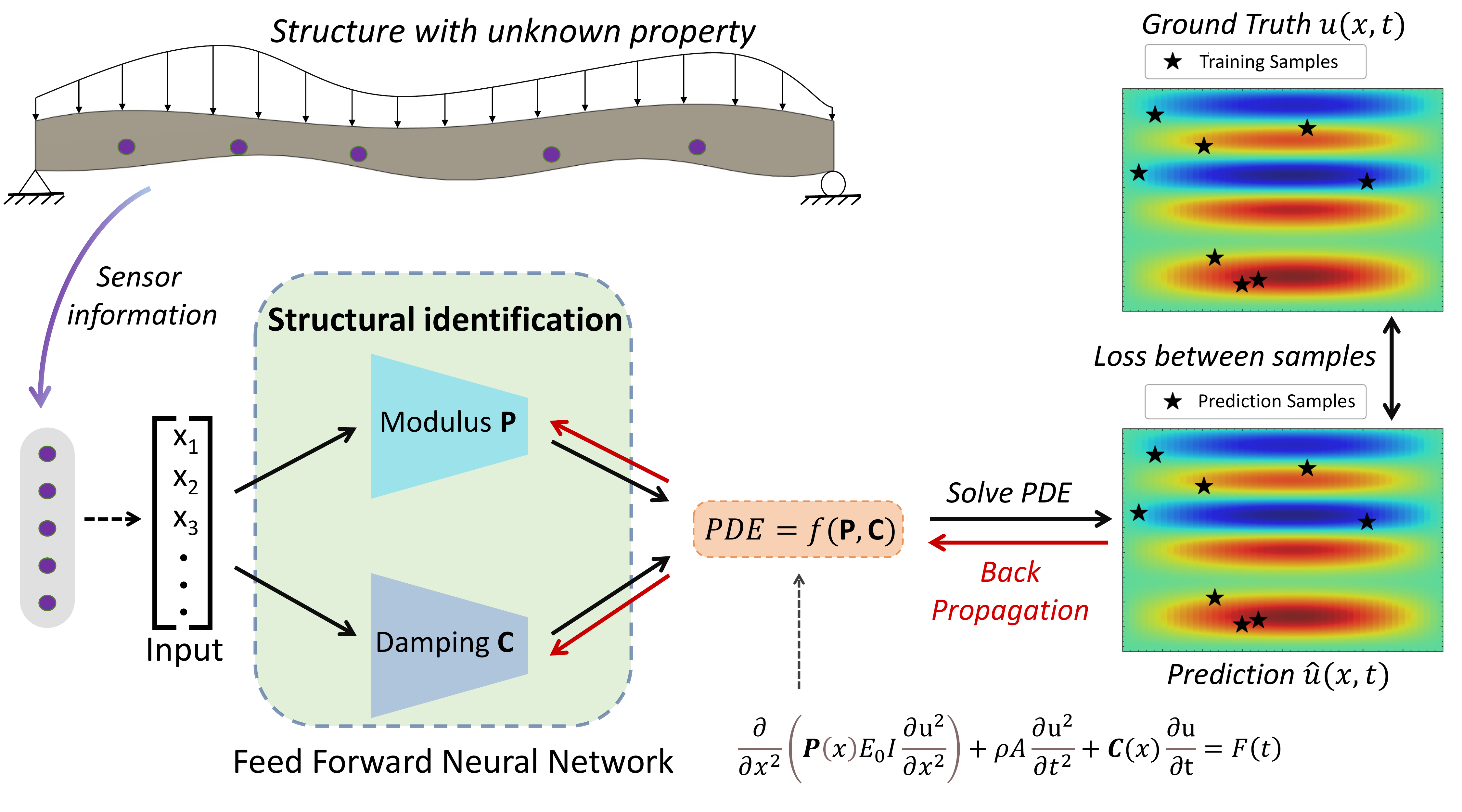}
    \caption{\textbf{Overview: } We consider structures whose dynamics are governed by a known partial differential equation (PDE), but with unknown parameters that potentially vary in both space and time. These unknown parameters are modeled with neural networks, which are then embedded within the PDE. In this illustration, the unknown parameters, modulus $P$ and damping $C$, vary spatially. The network weights are learned by solving the PDE to obtain the structural response (deflection in this case) and propagating the error between the predicted response and the measured ground truth response through the PDE solve and the neural networks. \label{fig:overview}} 
\end{figure}

In this paper, we consider the class of non-linear structural problems with unknown spatially distributed parameters (see Fig.\ref{fig:overview} for an overview). The parameters correspond to geometric and material variations and energy dissipation mechanisms, which could be due to damping or other system imperfections that are not typically captured in designs. As an instance of this problem class, we consider forced vibration responses in beams with spatially varying parameters. The primary challenges in such problems arise from the spatially variable nature of the properties and the distributed energy dissipation. This is typical for built civil structures, where energy dissipation and other hard-to-model phenomena physically drive the dynamic response behavior. In addition, it is very common to have structural systems with unknown strength distributions, which can be driven by geometric non-linearities or indiscernible/hidden material weaknesses. Finally, a typical challenge in structural systems is the rarity of measured data, especially for extreme loading cases.

We propose a framework, dubbed \ourmethod{}, for nonlinear dynamic system identification that allows us to discover the unknown parameters of partial  differential equations from measured sensing data. The developed model performance is compared to conventional PINN methods and direct regression models. Upon estimating the unknown system parameters, we apply them to the differential model and efficiently prognosticate the time evolution of the structural response. We also investigate the performance of \ourmethod{} under a limited training data regime across different input beam loading conditions. This replicates the expected challenges in monitoring real structures with limited sensors and sampling capabilities.

\ourmethod{} contributes to the fields of NeuralPDEs, structural identification, and health monitoring:
\begin{enumerate}
  \item \ourmethod{} allows us to learn unknown parameters of fundamental governing dynamics of structural systems expressed in the form of PDEs.
  \item We demonstrate the utility of \ourmethod{} by modeling the vibrations of nonlinear beams with unknown parameters. Experimental results demonstrate that \ourmethod{} achieves two-to-three orders of magnitude lower error in predicting displacement distributions in comparison to PINN-based baselines.
  \item We also demonstrate the utility of \ourmethod{} in temporally extrapolating displacement distribution predictions well beyond the training data measurements. Experimental results demonstrate that \ourmethod{} achieves four-to-five orders of magnitude lower error compared to PINN-based baselines.
\end{enumerate}

\section{Related Work}
Significant efforts have been directed toward physics-driven discovery or approximation of governing equations~\cite{ghorbani2020hybrid,maurya2020incorporating,lai2021structural}. Such studies have further been amplified by the rapid development of advanced sensing techniques and machine learning methods~\cite{hasni2017new,hasni2017self,konkanov2020environment,salehi2021comprehensive}. Most of the work to date has mainly focused on ordinary differential equation systems \cite{lai2021structural,zhang2019anodev2}. Neural ODEs~\cite{chen2018neural} have been widely adopted due to their capacity to learn and capture the governing dynamic behavior from directly collected measurements \cite{aliee2021beyond,zhang2019anodev2,rackauckas2020universal}. They represent a significant step above the direct fitting of a relation between input and output variables. In structural engineering applications, Neural ODEs generally approximate the time derivative of the main physical attribute through a neural network. 

More recently, data-driven discovery algorithms for the estimation of parameters in differential equations are introduced. These methods typically referred to as physics-informed neural networks (PINNs) include differential equations, constitutive equations, and initial and boundary conditions in the loss function of the neural network and adopt automatic differentiation to compute derivatives of the network parameters~\cite{krishnapriyan2021characterizing,raissi2019physics}. Variational Autoencoders were also learned to build baseline behavioral models, which were then used to detect and localize anomalies~\cite{li2022method}. Many other applications have employed Neural ODE for dynamic structure parameter identification in both linear and nonlinear cases~\cite{lai2021structural,aliee2021beyond,zhang2019anodev2}. On the other hand, few studies have explored Neural PDEs in other fields such as message passing \cite{brandstetter2022message}, weather and ocean wave data \cite{dulny2021neuralpde}, and fluid dynamics \cite{brandstetter2022lie}. In \cite{horie2022physics}, a Graph Neural Network was used to solve flow phenomena, and a NeuralPDE solver package was developed in Julia \cite{zubov2021neuralpde} based on PINN.

\section{Structural Problem – PDE derivation }
\subsection{Problem description}
Many physical processes in engineering can be described as fourth-order time-dependent partial differential problems. Examples include the Cahn-Hilliard type equations in Chemical Engineering, the Boussinesq equation in geotechnical engineering, the biharmonic systems in continuum mechanics, the Kuramoto-Sivashinsky equation in diffusion systems~\cite{modebei2020numerical} and the Euler-Bernoulli equation considered as an example case study in this paper. The Euler-Bernoulli beam equation is widely used in civil engineering to estimate the strength and deflection of beam structures. The dynamic beam response is defined by:

\begin{equation}
F(t)=\frac{\partial^2 }{\partial x^2} \biggl(P(x)E_0I\frac{\partial^2 u}{\partial x^2}\biggl) + \rho A\frac{\partial^2 u}{\partial t^2} + C(x) \frac{\partial u}{\partial t}
\end{equation}
\noindent where $u(x,t)$ is the displacement as a function of space and time. $P(x)$ and $E_0$ are the modulus coefficient and the reference modulus value of the beam, $I, \rho$, and $A$ are refereed to the beam geometry and density. $F$ is the distributed force applied to the beam. $C(x)$ represents damping, which is related to energy dissipation in the structure. In this paper, we restrict ourselves only to spatial variation of the beam’s properties and leave the most generalized case with variations in space and time of all variables for a future study. 

\begin{figure}[!h]
\centering
\includegraphics[height=2.9cm]{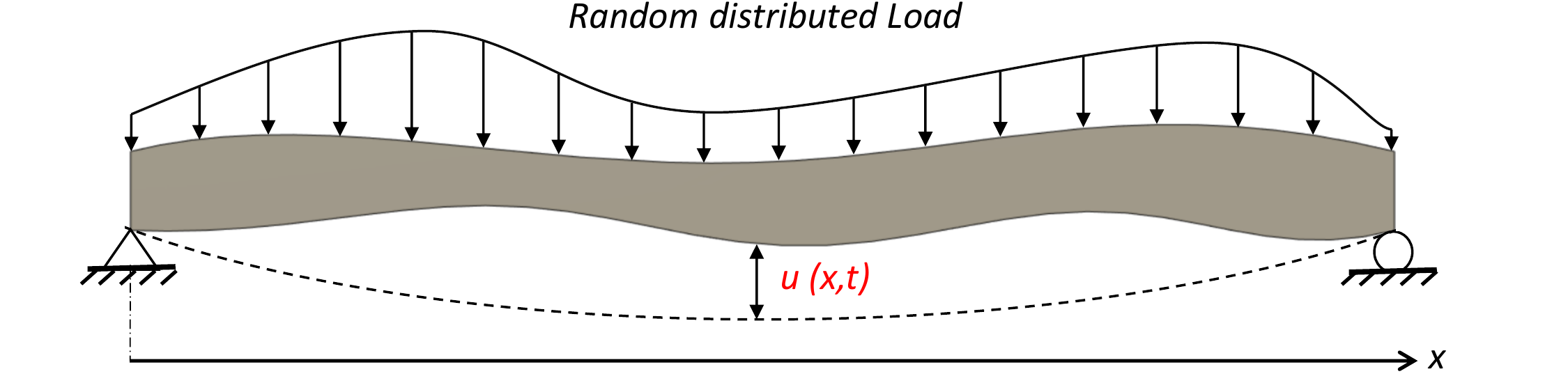}
\caption{Simply supported dynamic beam bending problem. Dynamic load can be applied to the structure with its values changing in time. The geometry, modulus, and other properties of the beam can also vary spatially with $x$. The deflection of the beam is defined as $u(x,t)$. \label{fig:beam}
}
\end{figure}

The fourth-order derivative of the spatial variable and the second-order derivative of time describes the relation between the beam deflection and the load on the beam \cite{akinpelu2012response}. Figure~\ref{fig:beam}. shows an illustration of the beam problem considered here, with the deflection $u(x,t)$ as the physical response of interest. The problem can also be formulated as a function of moments, stresses, or strains. The deflection formulation presents the highest order differentiation in the PDE. This was selected to allow for flexibility of the solution to be extended to other applications beyond structural engineering.  

To accurately represent the behavior of a structural component, its properties need to be identified. Though the beam geometry is straightforward to measure, the material property and damping coefficient are hard to estimate. The beam reference modulus $E_0$ is expected to have an estimated range based on the choice of material (e.g., steel, aluminum, composites, etc.) but unforeseen weaknesses in the build conditions can introduce unexpected nonlinear behavior. One of the objectives of this work is to capture this indiscernible randomness from response measurements. In addition, as discussed above, the damping is unpredictable at the design stage and is usually calculated by experiments. For the simply supported beam problem, the boundary conditions are defined as:
\begin{equation}
\begin{dcases}
    u(x=0,t)=0\text{;}\quad\quad u(x=L,t)=0 \\
    \frac{\partial^2 u(x=0,t)}{\partial x^2}=0 \text{;}\quad \frac{\partial^2 u(x=L,t)}{\partial x^2}=0 
\end{dcases}
\end{equation}
\noindent where $L$ is the length of the beam. Initially, the beam is static and stable, so the initial conditions of the beam are:
\begin{equation}
\begin{dcases}
    u(x,t=0)=0 \\
    \frac{\partial u(x,t=0)}{\partial t}=0
\end{dcases}
\end{equation}

\section{\ourmethod{}}
\subsection{Discretization of Space}
To tackle this high-order PDE efficiently, a numerical approach based on the method of lines is employed to discretize the spatial dimensions of the PDE. Then the system is solved as a system of ordinary differential equations (ODEs). The implemented discretization for the spatial derivatives of different orders are expressed as:

\begin{equation}
A_4^*u/\Delta x^4 = \frac{\partial^4 u}{\partial x^4}\text{;}\quad A_3^*u/\Delta x^3 = \frac{\partial^3 u}{\partial x^3}\text{;}\quad A_2^*u/\Delta x^2 = \frac{\partial^2 u}{\partial x^2}
\end{equation}
\noindent where in the fourth order discretization, $A_4^*$ is a $N \times N$ modified band matrix (based on the boundary conditions), and the size depends on the number of elements used for the space discretization, and $\Delta x$ is the distance between the adjacent elements discretized in the spatial domain. A similar principle is applied for other order derivatives.

\subsection{The proposed \ourmethod{} Schematic}
A pictorial schematic of \ourmethod{} is shown in Fig.~\ref{fig:overview}. The Julia differential equation package~\cite{rackauckas2020universal}  allows for very efficient computation of the gradient from the ODE solver. This makes it feasible to be used for neural network backpropagation. Thus, the ODE solver can be considered as a neural network layer after defining the ODE problem with the required fields of initial conditions, time span, and any extra parameters. Inputs to this layer can either be output from the previous network layers or directly from the training data.

The network in \ourmethod{} for the beam problem takes as input the location of the deformation sensors installed on the structure for continuous monitoring of its response. A series of dense layers are implemented to produce the output, which are the parameters that represent the structural characteristics. The parameters are re-inserted into the pre-defined ODE to obtain the final output, i.e, the structure's dynamic response. The loss is determined by the difference between the dynamic responses predicted by \ourmethod{} and those measured by the sensors (ground truth).

\subsection{Training data generation}
\begin{figure}[!ht]
\centering
\includegraphics[width=0.9\textwidth]{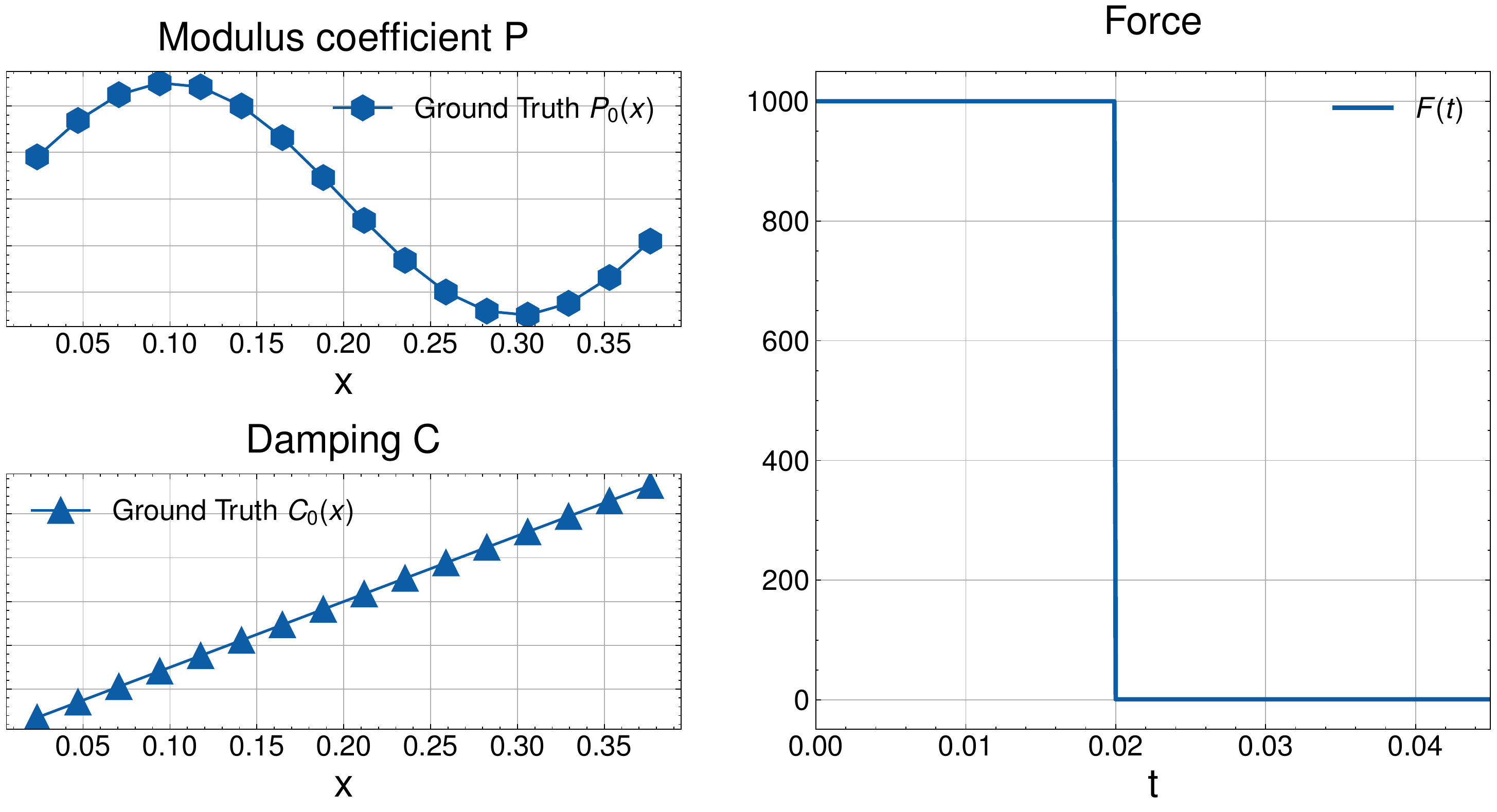}
\caption{Pre-defined structural properties and resultant dynamic response. Structural parameters $P$ and $C$ are defined as a sinusoidal and a ramp function. Force is applied as a step function of 1000 N and reduced to zero after 0.02s.}
\label{fig:fig3}
\end{figure}

For experimental considerations in future lab testing, we simulate in this case a beam with length, width, and thickness respectively of $40cm$, $5cm$, and $0.5cm$. The density $\rho$ is $2700kg/m^3$ (aluminum as base material). The force $F(t)$ is defined as a nonlinear temporal function. Considering the possible cases of polynomial or harmonic material properties variations as an example~\cite{salem2021functionally}, we integrate the beam with a nonlinear modulus $E(x)$ as a sinusoidal function. We use a range for the modulus from $70GPa$ to $140GPa$ (again using aluminum as a base reference). The damping coefficient $C(x)$ is modeled as a ramp function. The PDE can be rewritten and expressed as:
\begin{equation}
\begin{split}
F(t)=E_0I\Bigl(\frac{\partial^2 P(x)}{\partial x^2} \frac{\partial^2 u}{\partial x^2} + 2\frac{\partial P(x)}{\partial x} \frac{\partial^3 u}{\partial x^3}  + P(x)\frac{\partial^4 u}{\partial x^4}\Bigl) + \rho A\frac{\partial^2 u}{\partial t^2} + C(x) \frac{\partial u}{\partial t} \\
=E_0I\Bigl(A_2^*P(x) A_2^*u + 2A_1^*u P(x)I A_3^*u + P(x)I A_4^*u\Bigl) + \rho A\frac{\partial^2 u}{\partial t^2} + C(x) \frac{\partial u}{\partial t}
\label{eq:eq_pde}
\end{split}
\end{equation}

\begin{equation}
F(t) = \left\{
        \begin{array}{ll}
            1000 & \quad t \leq 0.02s \\
            0 & \quad t > 0.02s
        \end{array}
    \right.
\end{equation}
\noindent where the estimated modulus reference $E_0$ is $70GPa$, and $P(x)$ and $C(x)$ are modulus coefficient and damping that can vary spatially with $x$. The pre-defined parameters $P_0 (x)$ and $C_0 (x)$ are shown in Fig.\ref{fig:fig3}.

The PDE presented in \eqref{eq:eq_pde} is solved via the differential equation package in Julia. The RK4 solver method is selected for this high-order PDE. The time span was set to 0.045$s$ to have 3 complete oscillations of the bending response. The number of spatial elements and time steps are chosen as 16 and 160 respectively for balancing the training time cost and response resolution (capture the peak deflections). The deflections $u(x,t)$ are presented as a displacement distribution of size $16\times160$, from which ground truth data is obtained for training.

\subsection{Network architecture and training}
The network architecture is presented as a combination of multiple dense layers and an PDE-solver layer. The input to the network is the spatial coordinates $x$ for the measurements, and the network output is the prediction of the dynamic response $u(x,t)$. It is worth mentioning that the structural parameters $P$ and $C$ are produced from the multiple dense layers in separate networks, and the PDE layer takes those parameters to generate a response displacement distribution of size $16\times160$. The activation function for predicting the parameter $P$ is a linear scale of the sigmoid function so that the output can be in a reasonable range. For the prediction of parameter $C$, the network of the same architecture is used, but the last layer does not take any activation function since the range of the damping value is unknown.

\begin{figure}
\centering
\includegraphics[height=5.2cm]{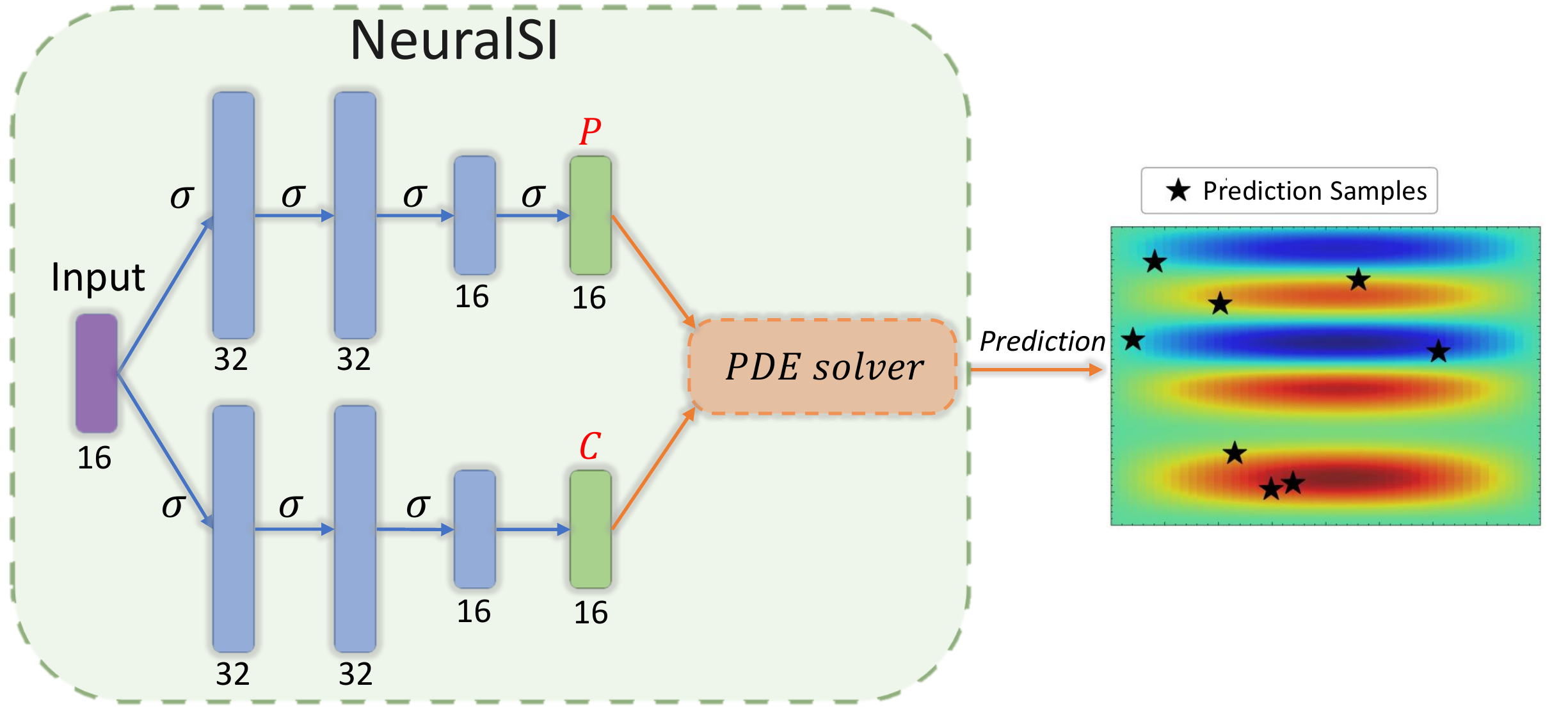}
    \caption{\ourmethod{} network architecture and training. The network has several dense layers and the output is split into $P$ and $C$. Those parameters are taken to the PDE solver for structural response prediction. Samples are taken randomly from the response for training the network.}
\label{fig:fig5}
\end{figure}

The modulus coefficient might be very high during training and lead to erroneous predictions with very high-frequency oscillations. So, we used minibatch training to escape local minima with a batch size of 16. The loss function is defined as the mean absolute error (MAE) between samples from the predicted and ground truth displacement distribution:
\begin{equation}
loss = \frac{1}{n}\sum_{i=1}^n \left|u-\hat{u}\right|
\end{equation}
where $n$ is the number of samples for training, $u$ and $\hat{u}$ are the values from true and prediction dynamic responses at different training points in the same minibatch.

Furthermore, inspired by the effectiveness of positional embeddings for representing spatial coordinates in transformers~\cite{vaswani2017attention}, we adopt the same as well as for the spatial input to the network. It is worth noting that the temporal information in the measurements is only used as an aid for mapping and matching the predictions with the ground truth. We use ADAMW~\cite{loshchilov2018fixing} as our optimizer, with a learning rate of 0.01.

\section{Results and Performance}
The evaluation of \ourmethod{} is divided into two parts. In the first part, we evaluate predictions of the parameters $P$ and $C$ from the trained neural network. We assume that each structure has a unique response. To determine how well the model is predicting the parameters, Fr\'{e}chet distance \cite{eiter1994computing} is employed to estimate the similarity between the ground truth and predicted functions. In this case, the predicted $P$ and $C$ are compared to the original $P_0$ and, $C_0$ respectively.

The second part of our evaluation is the prediction of the dynamic responses, which is achieved by solving the PDE using the predicted parameters. The metric to determine the performance of the prediction is the mean average error (MAE) between the predicted and ground truth displacement distribution. The prediction can be extrapolated by solving the PDE for a longer time span and compared with the extrapolated ground truth. The MAE is also calculated from the extrapolated data to examine the extrapolation ability of \ourmethod{}. Moreover, the dynamic response can be visualized on different elements separately (i.e., separate spatial locations $x$)  for a more fine-grained comparison of the extrapolation results.

\subsection{Results} 
We first trained and evaluated \ourmethod{} with different combinations of number and size of dense layers, percentage of data used for training, and minibatch size. The best results were achieved by taking a minibatch size of 16, training for a total of 20 epochs, and a learning rate of 0.001 (the first 10 epochs has a learning rate of 0.01).
\begin{figure}[!h]
\centering
\includegraphics[height=6cm]{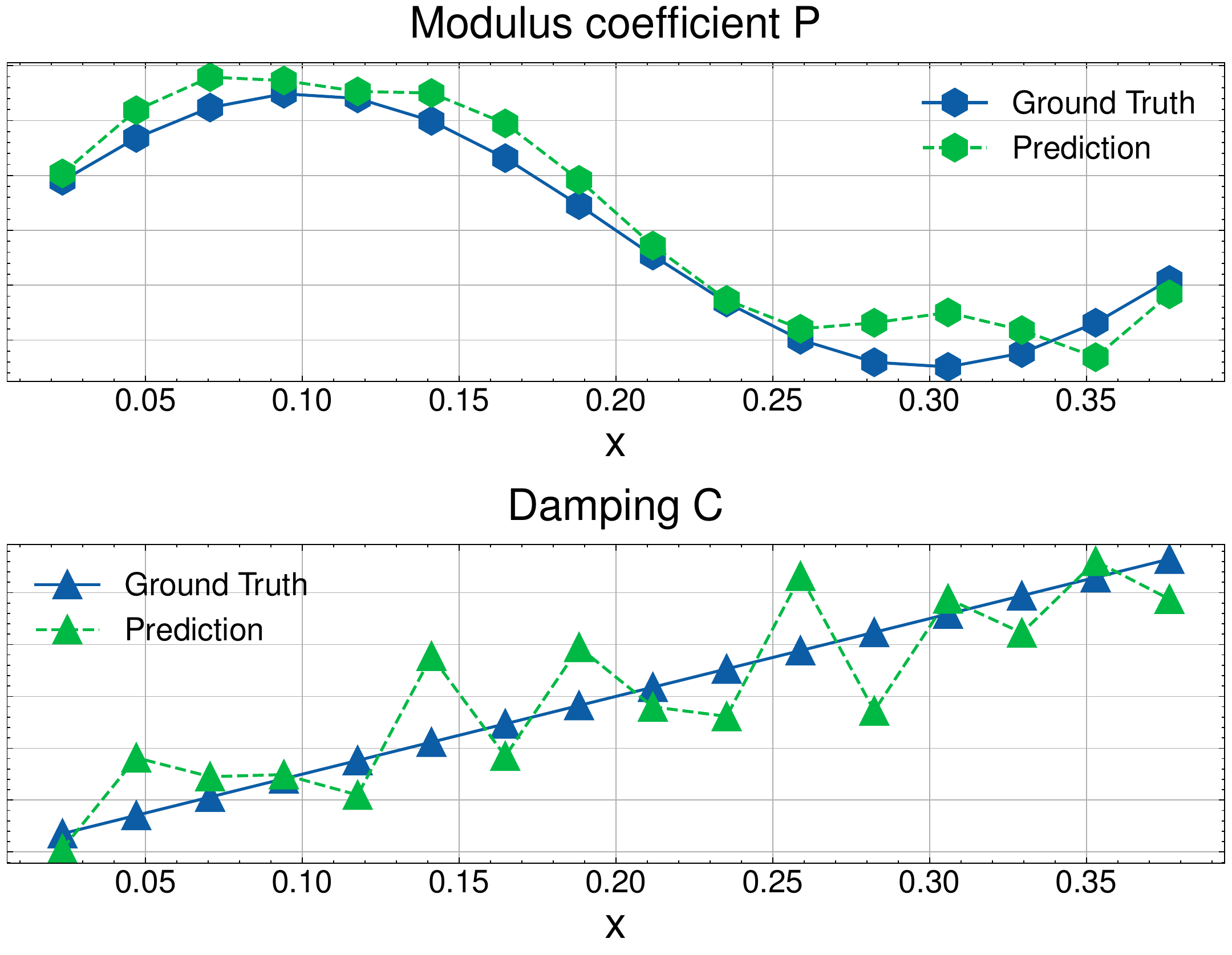}
\caption{Predicted beam parameters modulus coefficient (top) and damping (bottom). Observe that the modulus coefficient $P$ matches well with the sinusoidal ground truth, since the modulus dominates the magnitude of the response. The damping $C$ fluctuates as it is less sensitive than $P$, but the outputs still present a trend of increasing damping magnitude from the left end of the beam to the right end.\label{fig:V}}
\end{figure}

Figure~\ref{fig:V} shows the output of modulus coefficient $P$ and damping $C$ from \ourmethod{}. For the most part, the predictions match well with the target modulus and damping, respectively. Compared to the modulus coefficient $P$, the predicted damping $C$ has a larger error since it is less sensitive to the response. A small difference in damping magnitude will not affect the dynamic response as much as a change in the modulus parameter. However, the non-linearity of the modulus and damping are predicted accurately, and it is easy to identify whether the system is under-damped or over-damped based on the predicted damping parameters.
\begin{figure}[!ht]
\centering
\begin{subfigure}{\textwidth}
    \centering
    \includegraphics[height=3.1cm]{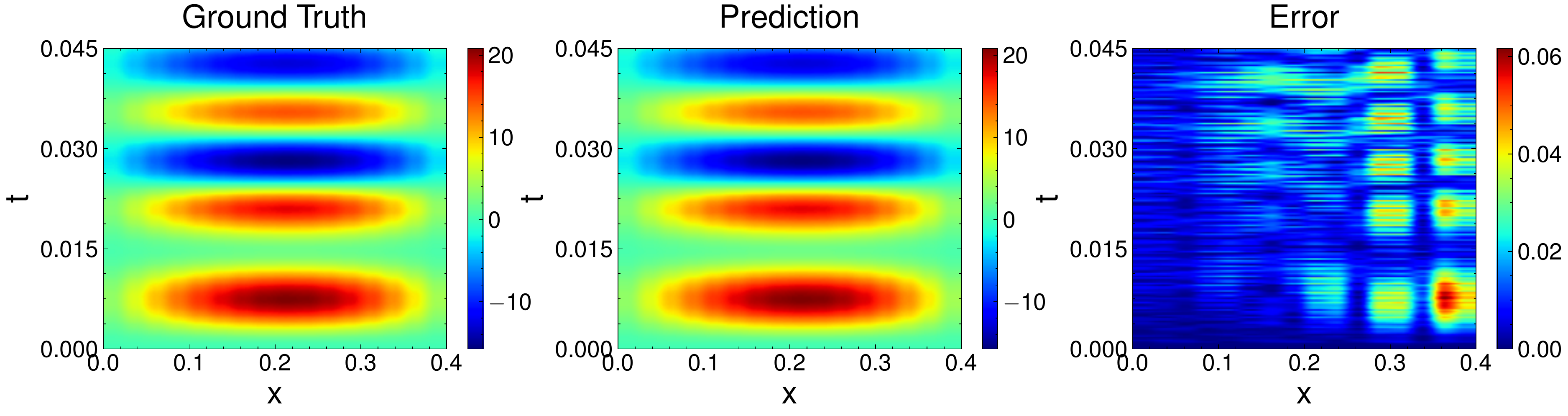}
\end{subfigure}
\begin{subfigure}{\textwidth}
    \centering
    \includegraphics[height=3.1cm]{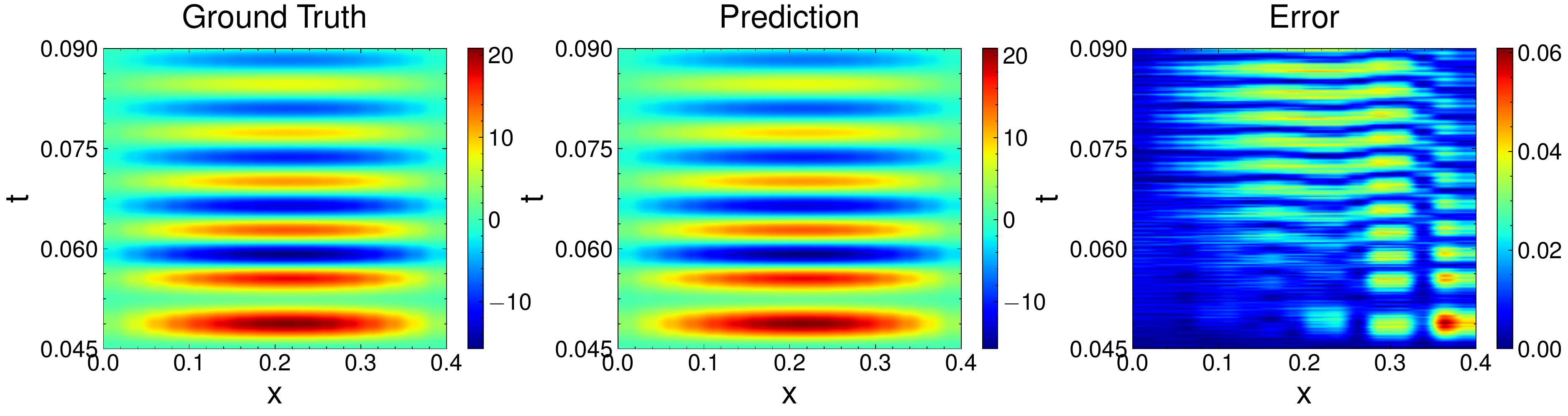}
\end{subfigure}
\caption{\ourmethod{} predictions. The interpolation results (top row) are calculated from 0 to 0.045s and temporal extrapolation results (bottom row) are from 0.045s to 0.09s. Peak error is only around 0.3\% of the peak value from the ground truth, and the error magnitude remains the same for extrapolation.\label{fig:heatmap}}
\end{figure}

\begin{figure}[h]
\centering
\includegraphics[height=3.9cm]{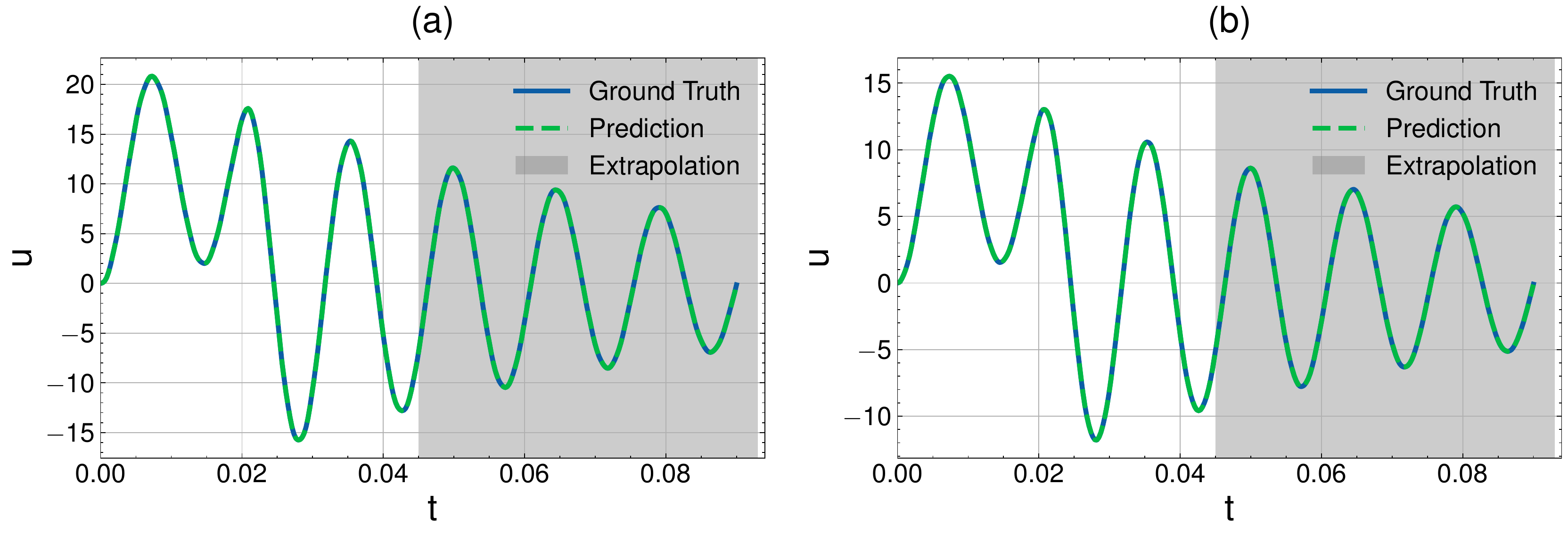}
\caption{Elemental response, spatial elements from the beam are selected to examine the temporal response. The ground truth and prediction responses are matching perfectly. (a) element at beam midspan; (b) element at quarter length of the beam.\label{fig:element}}
\end{figure}

Figure~\ref{fig:heatmap} visualizes the ground truth and predicted dynamic displacement response, along with the error between the two. We observe that the maximum peak-peak value in the displacement error is only 0.3\% of the ground truth. We also consider the ability of \ourmethod{} to extrapolate and display the dynamic response by doubling the prediction time span. It is worth mentioning that the peak error in temporal extrapolation does not increase much compared to the peak error in temporal interpolation. The extrapolation results are also examined at different elements from different locations. Figure~\ref{fig:element} presents the response at the beam midspan and at quarter length. There are no observed discrepancies between the ground truth and the predicted response.

\subsection{Hyperparameter Investigation}
Based on the parameters chosen above, we tested the effect of number of dense layers, training sample ratio and minibatch size on the parameter identification and prediction of dynamic responses. 
\begin{figure}[!ht]
\centering
\begin{subfigure}{\textwidth}
    \centering
    \includegraphics[width=\textwidth]{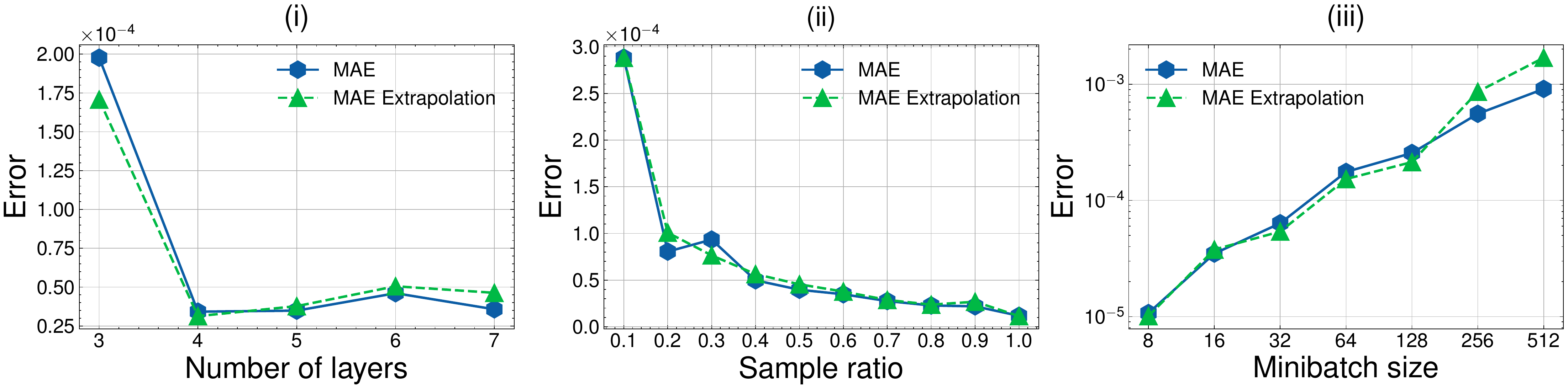}
    \caption{MAE (i) Number of layers; (ii) Sample ratio; (iii) Minibatch size \label{fig:MAE}}
\end{subfigure}
\begin{subfigure}{\textwidth}
    \centering
    \includegraphics[width=\textwidth]{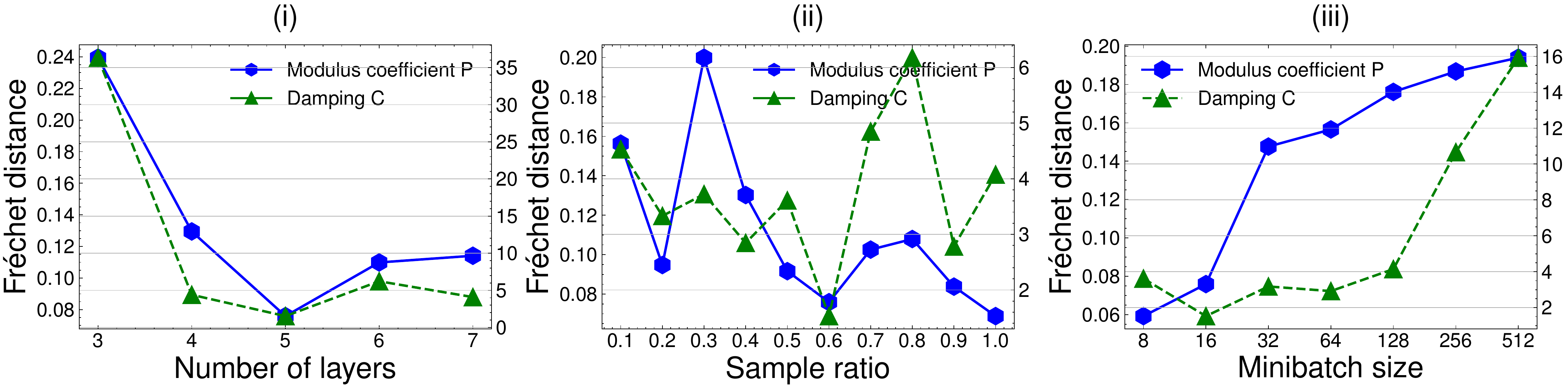}
    \caption{Fr\'{e}chet distance (i) Number of layers; (ii) Sample ratio; (iii) Minibatch size \label{fig:distance}}
\end{subfigure}
\caption{Hyperparameter performance. A sufficient number of layers, more training samples, and small minibatch size will produce a good combination of hyperparameters and loss MAE (top row). The Fr\'{e}chet distances (bottom row) are calculated for $P$ and $C$ respectively. The fluctuation of Fr\'{e}chet distance for different sample ratio is because the values are relatively small.\label{fig:hyperparameter}}
\end{figure}

\subsubsection{Number of layers:}
The number of layers is varied by consecutively adding an extra layer with 32 hidden units right after the input. From Fig.~\ref{fig:hyperparameter}, the performance of the network is affected if the number of layers is below 4. This is explained by the fact that the network does not have sufficient capacity to precisely estimate the unknown structural parameters. It is noted that the size of the input and output are determined by the minibatch size and the number of elements used for discretization. A higher input or output size will automatically require a bigger network to improve prediction accuracy. Additionally, the Fr\'{e}chet distance decreases as the size of the neural network increases, which demonstrates that the prediction of beam parameters is more accurate.

\subsubsection{Sample ratio:}
The number of training samples plays an important role in the model and in real in-field deployment scenarios. The number and the efficiency of sensor arrangements will be directly related to the number of samples required for accurately estimating the unknown parameters. It is expected that a reduced amount of data is sufficient to train the model given the strong domain knowledge (in the form of PDE) leveraged by \ourmethod{}. From Fig.~\ref{fig:hyperparameter}, when 20\% of the ground truth displacement samples are used for training, the loss drops noticeably. With an increased amount of training data, the network performance can still be improved. Furthermore, observe that there is a slight effect of data overfitting when using the full amount of data for training. The Fr\'{e}chet distance of damping is not stable since our loss function optimizes for accurately predicting the dynamic deflection response, instead of directly predicting the parameters. As such, the same error could be obtained through different combinations of those parameters.

\subsubsection{Minibatch size:}
The minibatch size plays an important role in the efficiency of the training process and the performance of the estimated parameters. It is worth mentioning that a smaller minibatch size helps escape local minima and reduces errors. However, this induces a higher number of iterations for a single epoch, which is computationally expensive. From Fig.~\ref{fig:hyperparameter} we observe that both the MAE error and the Fr\'{e}chet distance are relatively low when the minibatch size is smaller than 32. 

\section{Comparison of \ourmethod{} with a direct response mapping Deep Neural Network and a PINN}

The \ourmethod{} framework is compared with traditional deep neural networks (DNN) and PINN methods. The tested DNN has 5 dense layers and a Tanh activation. The inputs are the spatial and temporal coordinates $x$ and $t$, respectively, of the displacement response, and the output is the beam deflection $u(x,t)$ at that spatio-temporal position. The optimizer is LBFGS and the learning rate is 1.0. With a random choice of 20\% samples, the loss stabilizes after 500 epochs.

The PINN method is defined with a similar strategy to existing solutions~\cite{krishnapriyan2021characterizing,raissi2019physics}. The Neural network consists of 5 dense layers with Tanh activation function. The loss is defined as a weighted aggregate of the boundary condition loss (second derivative of input $x$ at the boundaries), governing equation loss (fourth-order derivative of $x$ and second-order derivative of the $t$), and loss between the prediction and ground truth displacement response. We used LBFGS as the optimizer with a learning rate of 1.0. The training was executed for 3700 epochs.

\begin{figure}[!h]
\centering
\begin{subfigure}{\textwidth}
    \centering
    \includegraphics[width=\textwidth]{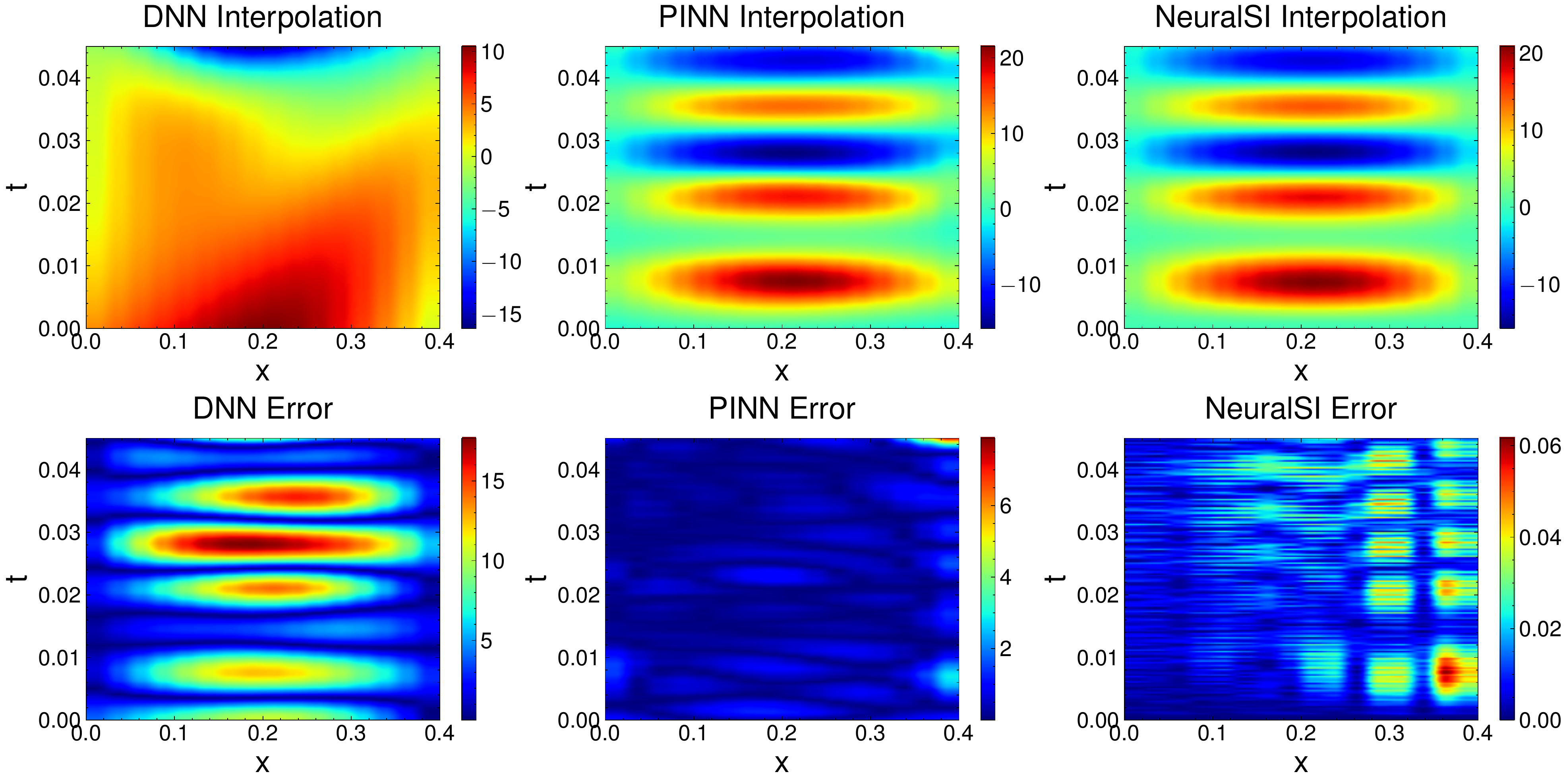}    
\end{subfigure}
\begin{subfigure}{\textwidth}
    \centering
    \includegraphics[width=\textwidth]{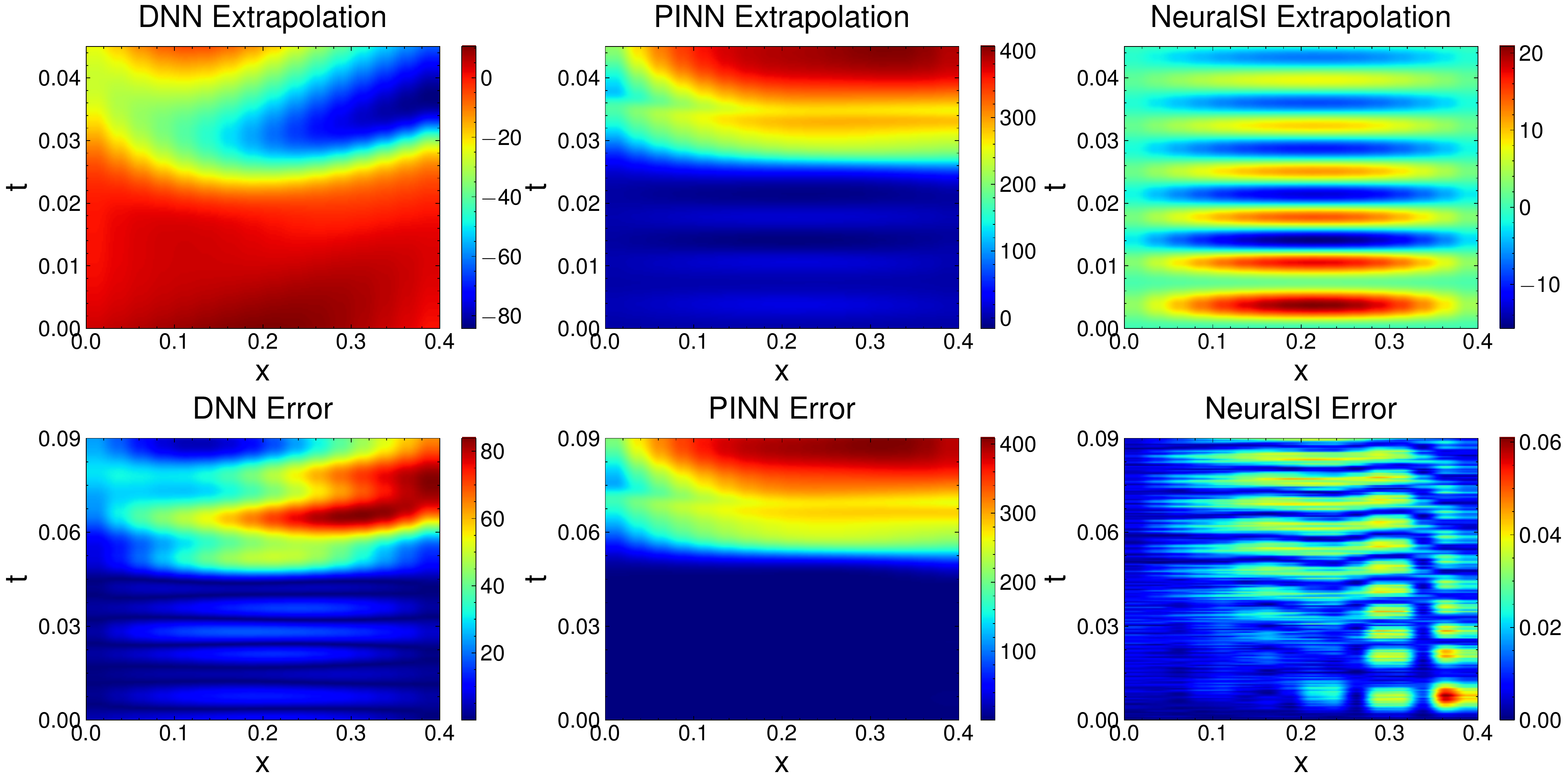}
\end{subfigure}
\caption{Spatio-temporal displacement distribution predictions and comparisons between DNN, PINN and \ourmethod{} for both interpolation (top) and extrapolation (bottom). The DNN method fails to learn the interpolation response, while the PINN can predict most of the responses correctly, with only a few errors at the corners of the displacement response. Predictions from \ourmethod{} have two orders of magnitude lower error in comparison to PINN. With the learned structural parameters, \ourmethod{} maintains the same magnitude of error in extrapolation results. Both DNN and PINN completely fail at extrapolation and lead to considerable errors.\label{fig:heatmaps}}
\end{figure}

\begin{figure}[!h]
\centering
\includegraphics[height=3.8cm]{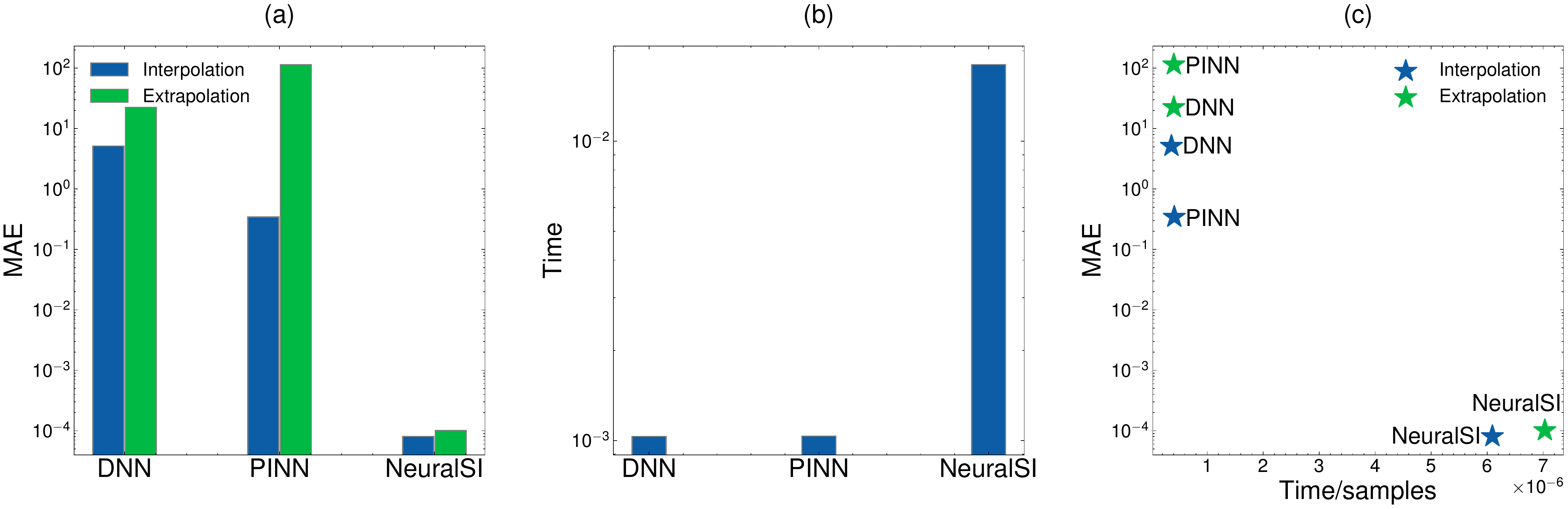}
\caption{Performance comparison between DNN, PINN, and \ourmethod{} for both interpolation and extrapolation (a) MAE, (b) Inference time, and (c) Trade-off between MAE and inference time. \ourmethod{} offers significantly lower error while being as expensive as solving the original PDE, thus offering a more accurate solution when the computational cost is affordable. \ourmethod{} obtains the extrapolation results by solving the whole time domain starting from $t=0$, while DNN and PINN methods directly take the spatio-temporal information and solve for extrapolation.  \label{fig:errors}}
\end{figure}

The prediction of the dynamic deformation responses for the two baseline methods and \ourmethod{} and the corresponding displacement distribution errors are shown in Fig.~\ref{fig:heatmaps}. In \ourmethod{}, we used ImplicitEulerExtrapolation solver for a $4\times$ faster inference. We further optimized the PDE function with ModelingToolkit \cite{ma2021modelingtoolkit}, which provides another $10\times$ speedup, for a total of $40\times$ speedup over the RK4 solver used for training. Due to a limited amount of data for training, the DNN fails to predict the response. With extra information from the boundary conditions and equation, the PINN method results in an MAE loss of 0.344, and the prediction fits the true displacement distribution well. Most of the values in the displacement distribution error are small, except for some corners. But both methods fail to extrapolate the structural behavior temporally. The extrapolation of DNN predictions produces large discrepancies compared to the ground truth. Similarly, the PINN method fails to match the \ourmethod{} performance, while fairing much better than the predictions from the DNN, as expected due to the added domain knowledge. The MAE errors were computed and compared with the proposed method trained with 20\% data as shown in Fig.~\ref{fig:errors}.

\section{Conclusion}

In this paper, we proposed \ourmethod{}, a framework that can be employed for structural parameter identification in nonlinear dynamic systems. Our solution models the unknown parameters via a learnable neural network and embeds it within a partial differential equation. The network is trained by minimizing the errors between predicted dynamic responses and ground truth measurement data. A major advantage of the method is its versatility and flexibility; thus, it can be successfully extended to any PDEs with high-order derivatives and nonlinear characteristics. The trained model can be used to either explore structural behavior under different initial conditions and loading scenarios, which is vital for structural modeling or to determine high-accuracy extrapolation, also essential in systems’ response prognosis. An example beam vibration study case was analyzed to demonstrate the capabilities of the framework. The estimated structural parameters and the dynamic response variations match well with the ground truth (MAE of $10^{-4}$). The performance of \ourmethod{} is also shown to outperform direct regression significantly through deep neural networks and PINN methods by three to five orders of magnitude.

\subsubsection{Acknowledgements:} This research was funded in part by the National Science Foundation grant CNS 1645783

\bibliographystyle{splncs04}
\bibliography{egbib}

\begin{thebibliography}{10}
\providecommand{\url}[1]{\texttt{#1}}
\providecommand{\urlprefix}{URL }
\providecommand{\doi}[1]{https://doi.org/#1}

\bibitem{akinpelu2012response}
Akinpelu, F.O.: The response of viscously damped euler-bernoulli beam to
  uniform partially distributed moving loads. Applied Mathematics
  \textbf{3}(3),  199--204 (2012)

\bibitem{aliee2021beyond}
Aliee, H., Theis, F.J., Kilbertus, N.: Beyond predictions in neural odes:
  Identification and interventions. arXiv preprint arXiv:2106.12430  (2021)

\bibitem{bagheri2018structural}
Bagheri, A., Ozbulut, O.E., Harris, D.K.: Structural system identification
  based on variational mode decomposition. Journal of Sound and Vibration
  \textbf{417},  182--197 (2018)

\bibitem{banerjee2010self}
Banerjee, B., Roy, D., Vasu, R.: Self-regularized pseudo time-marching schemes
  for structural system identification with static measurements. International
  Journal for Numerical Methods in Engineering  \textbf{82}(7),  896--916
  (2010)

\bibitem{brandstetter2022lie}
Brandstetter, J., Welling, M., Worrall, D.E.: Lie point symmetry data
  augmentation for neural pde solvers. arXiv preprint arXiv:2202.07643  (2022)

\bibitem{brandstetter2022message}
Brandstetter, J., Worrall, D., Welling, M.: Message passing neural pde solvers.
  arXiv preprint arXiv:2202.03376  (2022)

\bibitem{brincker2001modal}
Brincker, R., Zhang, L., Andersen, P.: Modal identification of output-only
  systems using frequency domain decomposition. Smart Materials and Structures
  \textbf{10}(3), ~441 (2001)

\bibitem{brunton2016discovering}
Brunton, S.L., Proctor, J.L., Kutz, J.N.: Discovering governing equations from
  data by sparse identification of nonlinear dynamical systems. Proceedings of
  the National Academy of Sciences  \textbf{113}(15),  3932--3937 (2016)

\bibitem{chen2018neural}
Chen, R.T., Rubanova, Y., Bettencourt, J., Duvenaud, D.K.: Neural ordinary
  differential equations. Advances in Neural Information Processing Systems
  \textbf{31} (2018)

\bibitem{diez2016clustering}
Diez, A., Khoa, N.L.D., Makki~Alamdari, M., Wang, Y., Chen, F., Runcie, P.: A
  clustering approach for structural health monitoring on bridges. Journal of
  Civil Structural Health Monitoring  \textbf{6}(3),  429--445 (2016)

\bibitem{dulny2021neuralpde}
Dulny, A., Hotho, A., Krause, A.: Neuralpde: Modelling dynamical systems from
  data. arXiv preprint arXiv:2111.07671  (2021)

\bibitem{eiter1994computing}
Eiter, T., Mannila, H.: Computing discrete fr{\'e}chet distance  (1994)

\bibitem{entezami2020big}
Entezami, A., Sarmadi, H., Behkamal, B., Mariani, S.: Big data analytics and
  structural health monitoring: a statistical pattern recognition-based
  approach. Sensors  \textbf{20}(8), ~2328 (2020)

\bibitem{entezami2017structural}
Entezami, A., Shariatmadar, H., Sarmadi, H.: Structural damage detection by a
  new iterative regularization method and an improved sensitivity function.
  Journal of Sound and Vibration  \textbf{399},  285--307 (2017)

\bibitem{ghorbani2020hybrid}
Ghorbani, E., Buyukozturk, O., Cha, Y.J.: Hybrid output-only structural system
  identification using random decrement and kalman filter. Mechanical Systems
  and Signal Processing  \textbf{144},  106977 (2020)

\bibitem{hasni2017new}
Hasni, H., Alavi, A.H., Jiao, P., Lajnef, N., Chatti, K., Aono, K.,
  Chakrabartty, S.: A new approach for damage detection in asphalt concrete
  pavements using battery-free wireless sensors with non-constant injection
  rates. Measurement  \textbf{110},  217--229 (2017)

\bibitem{hasni2017self}
Hasni, H., Alavi, A.H., Lajnef, N., Abdelbarr, M., Masri, S.F., Chakrabartty,
  S.: Self-powered piezo-floating-gate sensors for health monitoring of steel
  plates. Engineering Structures  \textbf{148},  584--601 (2017)

\bibitem{horie2022physics}
Horie, M., Mitsume, N.: Physics-embedded neural networks: E (n)-equivariant
  graph neural pde solvers. arXiv preprint arXiv:2205.11912  (2022)

\bibitem{konkanov2020environment}
Konkanov, M., Salem, T., Jiao, P., Niyazbekova, R., Lajnef, N.:
  Environment-friendly, self-sensing concrete blended with byproduct wastes.
  Sensors  \textbf{20}(7), ~1925 (2020)

\bibitem{krishnapriyan2021characterizing}
Krishnapriyan, A., Gholami, A., Zhe, S., Kirby, R., Mahoney, M.W.:
  Characterizing possible failure modes in physics-informed neural networks.
  Advances in Neural Information Processing Systems  \textbf{34},  26548--26560
  (2021)

\bibitem{lai2021structural}
Lai, Z., Mylonas, C., Nagarajaiah, S., Chatzi, E.: Structural identification
  with physics-informed neural ordinary differential equations. Journal of
  Sound and Vibration  \textbf{508},  116196 (2021)

\bibitem{lai2019sparse}
Lai, Z., Nagarajaiah, S.: Sparse structural system identification method for
  nonlinear dynamic systems with hysteresis/inelastic behavior. Mechanical
  Systems and Signal Processing  \textbf{117},  813--842 (2019)

\bibitem{li2022method}
Li, X., Salem, T., Bolandi, H., Boddeti, V., Lajnef, N.: Methods for the rapid
  detection of boundary condition variations in structural systems. American
  Society of Mechanical Engineers (2022)

\bibitem{loshchilov2018fixing}
Loshchilov, I., Hutter, F.: Fixing weight decay regularization in adam  (2018)

\bibitem{ma2021modelingtoolkit}
Ma, Y., Gowda, S., Anantharaman, R., Laughman, C., Shah, V., Rackauckas, C.:
  Modeling{T}oolkit: A composable graph transformation system for
  equation-based modeling. arXiv preprint arXiv:2103.05244  (2021)

\bibitem{maurya2020incorporating}
Maurya, D., Chinta, S., Sivaram, A., Rengaswamy, R.: Incorporating prior
  knowledge about structural constraints in model identification. arXiv
  preprint arXiv:2007.04030  (2020)

\bibitem{modebei2020numerical}
Modebei, M., Adeniyi, R., JATOR, S.: Numerical approximations of fourth-order
  pdes using block unification method. Journal of the Nigerian Mathematical
  Society  \textbf{39}(1),  47--68 (2020)

\bibitem{rackauckas2020universal}
Rackauckas, C., Ma, Y., Martensen, J., Warner, C., Zubov, K., Supekar, R.,
  Skinner, D., Ramadhan, A., Edelman, A.: Universal differential equations for
  scientific machine learning. arXiv preprint arXiv:2001.04385  (2020)

\bibitem{raissi2019physics}
Raissi, M., Perdikaris, P., Karniadakis, G.E.: Physics-informed neural
  networks: A deep learning framework for solving forward and inverse problems
  involving nonlinear partial differential equations. Journal of Computational
  Physics  \textbf{378},  686--707 (2019)

\bibitem{reynders2008reference}
Reynders, E., De~Roeck, G.: Reference-based combined deterministic--stochastic
  subspace identification for experimental and operational modal analysis.
  Mechanical Systems and Signal Processing  \textbf{22}(3),  617--637 (2008)

\bibitem{rezaiee2020sensitivity}
Rezaiee-Pajand, M., Entezami, A., Sarmadi, H.: A sensitivity-based finite
  element model updating based on unconstrained optimization problem and
  regularized solution methods. Structural Control and Health Monitoring
  \textbf{27}(5),  e2481 (2020)

\bibitem{salehi2021comprehensive}
Salehi, H., Burgue{\~n}o, R., Chakrabartty, S., Lajnef, N., Alavi, A.H.: A
  comprehensive review of self-powered sensors in civil infrastructure:
  State-of-the-art and future research trends. Engineering Structures
  \textbf{234},  111963 (2021)

\bibitem{salem2021functionally}
Salem, T., Jiao, P., Zaabar, I., Li, X., Zhu, R., Lajnef, N.: Functionally
  graded materials beams subjected to bilateral constraints: Structural
  instability and material topology. International Journal of Mechanical
  Sciences  \textbf{194},  106218 (2021)

\bibitem{sarmadi2020energy}
Sarmadi, H., Entezami, A., Daneshvar~Khorram, M.: Energy-based damage
  localization under ambient vibration and non-stationary signals by ensemble
  empirical mode decomposition and mahalanobis-squared distance. Journal of
  Vibration and Control  \textbf{26}(11-12),  1012--1027 (2020)

\bibitem{tuhta2019multi}
Tuhta, S., G{\"u}nday, F.: Multi input multi output system identification of
  concrete pavement using n4sid. International Journal of Interdisciplinary
  Innovative Research Development  \textbf{4}(1),  41--47 (2019)

\bibitem{vaswani2017attention}
Vaswani, A., Shazeer, N., Parmar, N., Uszkoreit, J., Jones, L., Gomez, A.N.,
  Kaiser, {\L}., Polosukhin, I.: Attention is all you need. Advances in Neural
  Information Processing Systems  \textbf{30} (2017)

\bibitem{yang2019automated}
Yang, X.M., Yi, T.H., Qu, C.X., Li, H.N., Liu, H.: Automated eigensystem
  realization algorithm for operational modal identification of bridge
  structures. Journal of Aerospace Engineering  \textbf{32}(2),  04018148
  (2019)

\bibitem{yin2017vibration}
Yin, T., Jiang, Q.H., Yuen, K.V.: Vibration-based damage detection for
  structural connections using incomplete modal data by bayesian approach and
  model reduction technique. Engineering Structures  \textbf{132},  260--277
  (2017)

\bibitem{yuen2004two}
Yuen, K.V., Au, S.K., Beck, J.L.: Two-stage structural health monitoring
  approach for phase i benchmark studies. Journal of Engineering Mechanics
  \textbf{130}(1),  16--33 (2004)

\bibitem{zhang2019anodev2}
Zhang, T., Yao, Z., Gholami, A., Gonzalez, J.E., Keutzer, K., Mahoney, M.W.,
  Biros, G.: {ANODEV2}: A coupled neural {ODE} framework. Advances in Neural
  Information Processing Systems  \textbf{32} (2019)

\bibitem{zhou2022semi}
Zhou, X., He, W., Zeng, Y., Zhang, Y.: A semi-analytical method for moving
  force identification of bridge structures based on the discrete cosine
  transform and fem. Mechanical Systems and Signal Processing  \textbf{180},
  109444 (2022)

\bibitem{zubov2021neuralpde}
Zubov, K., McCarthy, Z., Ma, Y., Calisto, F., Pagliarino, V., Azeglio, S.,
  Bottero, L., Luj{\'a}n, E., Sulzer, V., Bharambe, A., et~al.: Neuralpde:
  Automating physics-informed neural networks (pinns) with error
  approximations. arXiv preprint arXiv:2107.09443  (2021)

\end{thebibliography}
\end{document}